\pdfoutput=1

\documentclass[11pt]{article}

\usepackage{acl}

\usepackage{times}
\usepackage{latexsym}

\usepackage[T1]{fontenc}

\usepackage[utf8]{inputenc}

\usepackage{microtype}

\usepackage{inconsolata}

\usepackage{algorithm}
\usepackage{algorithmic}
\usepackage{amsmath}
\usepackage{bbm}
\usepackage{multicol}
\usepackage{hyperref}
\usepackage[nameinlink]{cleveref}

\usepackage{amssymb}
\usepackage{pifont}
\newcommand{\cmark}{\text{\ding{51}}}%
\newcommand{\xmark}{\text{\ding{55}}}%

\usepackage{color}
\usepackage{tabularray}

\usepackage{newunicodechar}

\usepackage{caption}
\usepackage{subcaption}

\usepackage{blindtext}
\usepackage{graphicx}
\usepackage{makecell}

\newcommand\blfootnote[1]{%
  \begingroup
  \renewcommand\thefootnote{}\footnote{#1}%
  \addtocounter{footnote}{-1}%
  \endgroup
}

\usepackage{booktabs}
\usepackage{multirow}
\usepackage[normalem]{ulem}
\useunder{\uline}{\ul}{}

\usepackage{lipsum}
\usepackage{tikz}
\usepackage{fancyvrb}
\usepackage{listings}
\usepackage{enumitem}
\usepackage{tcolorbox}
\usepackage{multicol}
\usepackage{siunitx}
\usepackage{xpatch}

\usepackage{lipsum}

\newcommand{\henry}[1]{\textcolor{red}{\bf\small [#1 --h]}}
\newcommand{\cornelia}[1]{\textcolor{blue}{\bf\small [#1 --cc]}}

\title{EIVEN: Efficient Implicit Attribute Value Extraction using \\ Multimodal LLM}

\author{Henry Peng Zou$^{\diamondsuit \dagger}$, Gavin Heqing Yu$^\spadesuit$, Ziwei Fan$^\spadesuit$, Dan Bu$^\spadesuit$, \\ \bf Han Liu$^{\heartsuit \dagger}$ Peng Dai$^\spadesuit$, Dongmei Jia$^\spadesuit$, Cornelia Caragea$^\diamondsuit$
\\  $^\spadesuit$Amazon \quad $^{\heartsuit}$Washington University in St. Louis \\ $^{\diamondsuit}$University of Illinois Chicago \\
\texttt{pzou3@uic.edu}
}

\begin{document}
\maketitle

\begin{abstract}

In e-commerce, accurately extracting product attribute values from multimodal data is crucial for improving user experience and operational efficiency of retailers. However, previous approaches to multimodal attribute value extraction often struggle with implicit attribute values embedded in images or text, rely heavily on extensive labeled data, and can easily confuse similar attribute values. To address these issues, we introduce EIVEN, a data- and parameter-efficient generative framework that pioneers the use of multimodal LLM for implicit attribute value extraction. EIVEN leverages the rich inherent knowledge of a pre-trained LLM and vision encoder to reduce reliance on labeled data. We also introduce a novel Learning-by-Comparison technique to reduce model confusion by enforcing attribute value comparison and difference identification. Additionally, we construct initial open-source datasets for multimodal implicit attribute value extraction. Our extensive experiments reveal that EIVEN significantly outperforms existing methods in extracting implicit attribute values while requiring less labeled data.\blfootnote{$^\dagger$Work done as an intern at Amazon.}
\end{abstract}




\section{Introduction}


Product attributes are crucial in e-commerce, aiding retailers in product representation, recommendation, and categorization, and assisting customers in product searching, comparison, and making informed purchasing decisions \citep{xu-etal-2019-scaling, yan-etal-2021-adatag, yang-etal-2023-mixpave, shinzato-etal-2023-unified}. Despite their importance, the accurate listing of these attributes remains a challenge. Sellers often fail to specify all relevant attribute values or list them incorrectly, leading to inefficiencies and potential customer dissatisfaction \cite{lin2021pam, khandelwal-etal-2023-large}. To address these issues, the task of Attribute Value Extraction (AVE) has emerged as a key area of research in e-commerce. AVE seeks to automate the extraction of attribute values from product profiles such as product titles, descriptions, and images \cite{zheng2018opentag, wang2020learning, wang-etal-2022-smartave}.

\begin{figure}[!t]
    \centering
    \includegraphics[width=1\columnwidth]{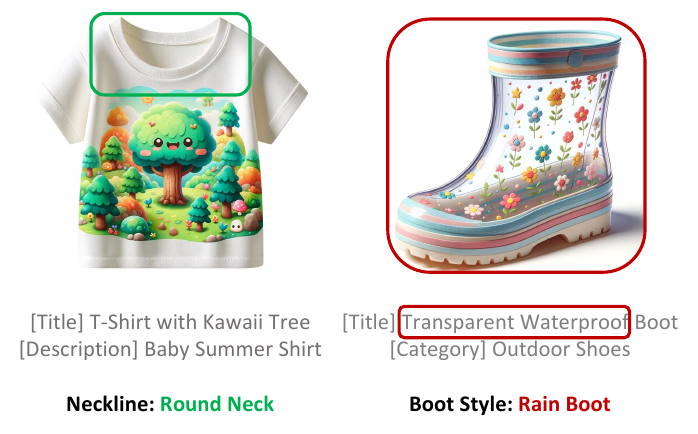} 
    \caption{Examples of implicit attribute values. The attribute value cannot be explicitly extracted as a part of product texts, but can inferred from the product image, text context or prior knowledge.}
    \label{fig:cover_example}
    \vspace{-20pt}
\end{figure}

Existing approaches for multimodal attribute value extraction can be broadly categorized into three categories: extractive, discriminative, and generative (more detailed discussion is provided in Appendix \ref{sec:discussion_multimodalAVE}).
Most extractive studies focus on extracting attribute values that are explicitly stated in product text data \cite{zhu-etal-2020-multimodal, yang2022mave, li-etal-2023-attgen, xu-etal-2023-towards}. However, in real-world scenarios, an attribute value that needs to be obtained may not appear as a subsequence of the product text, but can be inferred from the product image, implied text context or prior knowledge about this product type \cite{zhang-etal-2023-pay, khandelwal-etal-2023-large, blume-etal-2023-generative}. Take products in \textcolor{red}{Figure \ref{fig:cover_example}} for example. The value “round neck” of the “neckline” attribute does not appear in product textual information, but can be easily identified from its product image. Similarly, the value “rain boot” corresponding to the attribute “boot style" in the second product is not explicitly stated but is implicitly embedded in its textual context “transparent waterproof” and visual information. In addition, previous discriminative and generative approaches for multimodal AVE are highly data-hungry, requiring large amounts of labeled data for training but still perform poorly in extracting implicit attribute values \cite{zhang-etal-2023-pay, 9897323}. Furthermore, similar implicit attribute values are easily confused by the recent generative AVE model \cite{zhang-etal-2023-pay}.

To tackle these challenges, we introduce EIVEN, a data and parameter-efficient multimodal generative framework for multimodal implicit attribute value extraction. EIVEN utilizes the rich inherent knowledge of a pre-trained LLM and vision encoder to lessen reliance on extensive attribute-specific data. Additionally, to address the issue of model confusion caused by similar attribute values, we introduce a novel technique termed "Learning-by-Comparison". This approach feeds the model with pairs of instances that share the same attribute but potentially have different attribute values, forcing the model to compare and distinguish them.

Our contributions are summarized as follows:

\begin{itemize}[itemsep=0pt]
\vspace{-10pt}
\item To the best of our knowledge, we are the first work to explore multimodal LLM for the emerging real-world problem of implicit attribute value extraction.

\item We propose a novel Learning-by-Comparison technique to reduce model confusion among similar attribute values.

\item We construct initial open-source datasets for multimodal implicit AVE. \footnote{\textcolor{blue}{\url{https://github.com/HenryPengZou/EIVEN}}}


\item Extensive experiments show that our framework greatly outperforms recent multimodal AVE works, even with less labeled data.
\end{itemize}








\begin{figure*}[!th]
    \centering
    \includegraphics[width=\textwidth]{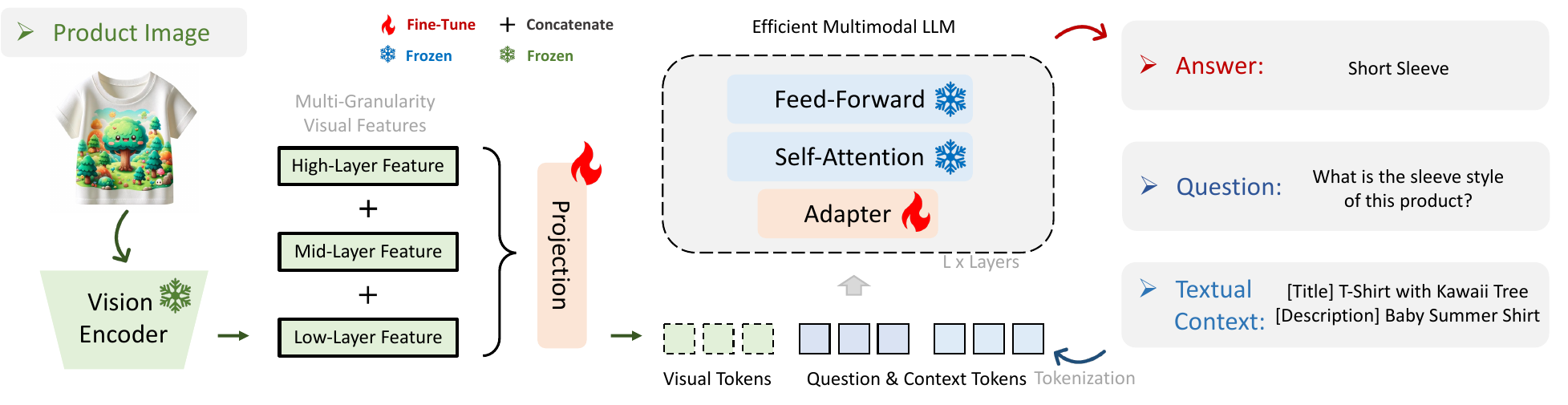} 
    \caption{Overview of our efficient multimodal LLM. We extract multi-granularity visual features from a frozen pre-trained vision encoder and use a learnable visual projection network to align their dimensions with text token embeddings. The obtained visual tokens and tokenized question and text context are fed to the LLM (LLaMA-7B) to generate the answer. We insert lightweight adapters into every layer of the LLM for parameter-efficient fine-tuning.}
    \label{fig:main_pipeline}
\end{figure*}

\begin{figure}[!th]
    \centering
    \includegraphics[width=1\columnwidth]{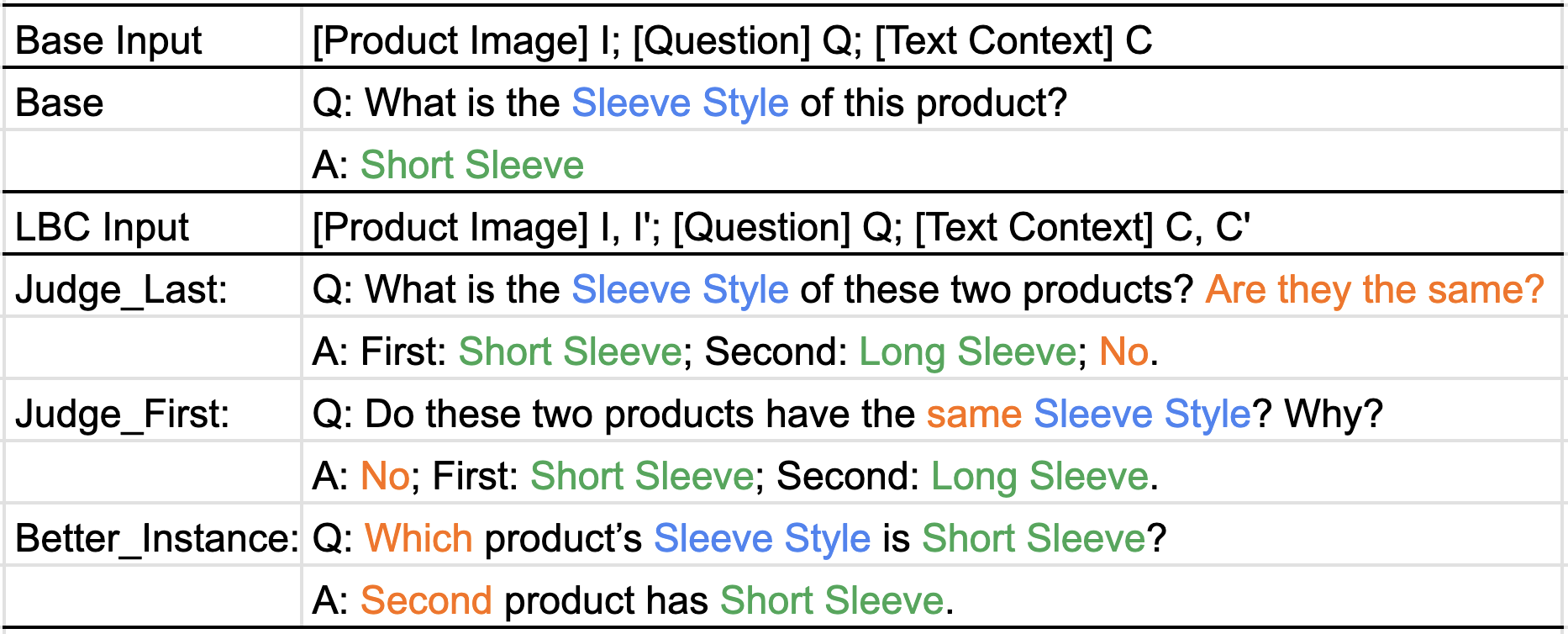} 
    \caption{Illustration of Learning-by-Comparison strategies. Our model is fed with pairs of product instances that share the same attribute but potentially different attribute values and asked to compare the values.}
    \label{fig:LBC}
\end{figure}

\section{EIVEN Framework}

\textbf{Problem Formulation.} Given a product's image and text context and a specified attribute, our goal is to extract the value for the corresponding attribute. Specifically, in our task of extracting implicit attribute values, the ground truth attribute value does not appear as a subsequence of the text context, but can be inferred from the product image, text context, or prior knowledge. In this work, we formulate the task of extracting implicit attribute values as the problem of generating answers given a question and product information. For example, the question could be "What is the Sleeve Style of this product?" and the generated answer could be "Short Sleeve" by inferring from the product's image and text context.

\textcolor{red}{Figure \ref{fig:main_pipeline}} presents an overview of our efficient multimodal LLM, and \textcolor{red}{Figure \ref{fig:LBC}} illustrates our Learning-by-Comparison strategies. Next, we explain our key components in detail.



\subsection{Image Embedding}

We leverage projected multi-granularity visual features to serve as the visual token input to our LLM model. Specifically, we extract visual features from the $[cls]$ token in every $M$ layer of the vision encoder and then concatenate them as:

$$I = \text{\,Concat\,}(\{I_k\}_{k=1}^{K})$$

\noindent where $K$ is the total number of extracted features, $I_k \in \mathbb{R}^{1 \times D}$ is the $k$-th extracted visual feature, and $I \in \mathbb{R}^{K \times D}$ is the overall multi-granularity image embedding.

Then, a simple visual projection network is used to adapt and transform the visual features to the same dimension as the text embedding of the LLM, which is denoted by:

$$I' = \sigma(I W_d + b_d)W_u + b_u$$

Here, $W_d \in \mathbb{R}^{D \times d_h}$ and $W_u \in \mathbb{R}^{d_h \times D_{text}}$ denote the weight matrices of the downsampling and upsampling layer, $b_d$ and $b_u$ are the bias terms, $\sigma$ is the SwiGLU activation function \cite{shazeer2020glu, luo2023cheap}. In this way, we empower the LLM to understand visual features at multiple levels of granularity, such as edges, textures, patterns, parts, and objects \cite{Ghiasi2022WhatDV, nguyen2019understanding}, which enables more effective extraction of attribute values.



\subsection{Efficient Multimodal LLM}
Previous generative works in multimodal implicit attribute value extraction \cite{zhang-etal-2023-pay, khandelwal-etal-2023-large} require large amounts of attribute-specific labeled data to achieve good performance. However, in the ever-evolving field of e-commerce, new products with unique attributes and values are constantly being introduced by different retailers and merchants. Gathering a large number of annotations for each new attribute is time-consuming and expensive \cite{yang-etal-2023-mixpave, Lai2021ASL, zou-caragea-2023-jointmatch, zou-etal-2023-decrisismb}. To reduce reliance on labeled data, we pioneer the exploration of leveraging pre-trained LLMs for the multimodal implicit AVE task. Trained on vast and diverse datasets, LLMs have demonstrated remarkable understanding, generative capabilities, and few-shot transfer learning ability \cite{Touvron2023LLaMAOA, liu2023visual, Wang2023KnowledgeGP, Tian2023GraphNP, Dong2023MuseChatAC, lai2024empowering}, making them a promising approach to be explored for implicit attribute value extraction.

However, LLMs typically comprise billions of parameters, rendering their full-scale fine-tuning both resource-demanding and inefficient. To address this, we resort to parameter-efficient fine-tuning strategies, which has been proven to achieve performance comparable to full fine-tuning but with substantially fewer trainable parameters \cite{hu-etal-2023-llm, houlsby2019parameter, luo2023cheap, Tian2024TinyLLMLA}. Specifically, we insert a lightweight adapter before every attention layer in our LLM. The mechanism of adapters is defined as:
$$h' = f_{\theta^u}  (\sigma(f_{\theta^d}(h))) + h$$
\noindent where $h, h'$ is the input and output of the adapter, $f_{\theta^d}(\cdot), f_{\theta^u}(\cdot)$ denotes for the downsampling and upsampling layers, $\sigma$ is an optional activation function depending on the choice of adapters. 


During training, we freeze all parameters in our LLM (LLaMA-7B \cite{Touvron2023LLaMAOA}) and the large image encoder, and only fine-tune these inserted lightweight adapters and the visual projection network.

Formally, given a product image embedding $I$, text context $C$, and an attribute-related question $Q$, the input of our multimodal LLM is denoted as $X = [I, Q, C]$. The overall training objective $\mathcal{L}$ of our multimodal LLM can be defined as:
$$\mathcal{L} = - \frac{1}{B} \sum_{i=1}^{B} \sum _{t=1}^{|R|} log \ p(R_t^i | X^i, R_{<t}^i; \theta_a, \theta_p)$$
\noindent where $B$ is the batch size, $R$ represents the ground-truth answer, $R_t$ is the $t$-th token of $R$, $R_{<t}$ represents the tokens before $R_t$, $\theta_a$ denotes all parameters of adapters in LLM, and $\theta_p$ denotes all parameters in the visual projection network.

In our training scheme, although we use LLM, thanks to these lightweight adapters, the number of trainable parameters can be kept at a very small scale, e.g., 2\textasciitilde5M. This greatly reduces the memory requirement and allows efficient training of EIVEN on the same single 32G V100 GPU as the previous work \cite{zhang-etal-2023-pay}, while achieving significantly better performance even with much less labeled data.

\begin{table*}[!tbh]
\resizebox{\textwidth}{!}{%
\begin{tabular}{@{}lc|ccc|ccc|ccc|c@{}}
\toprule
 & \multicolumn{1}{l}{} & \multicolumn{3}{c}{\textbf{Clothing}} & \multicolumn{3}{c}{\textbf{Footwear}} & \multicolumn{3}{c}{\textbf{General}} & \multicolumn{1}{l}{} \\ \midrule
\textbf{Method} & \textbf{Approach} & 10 & 100 & All & 10 & 100 & All & 10 & 100 & All & \textbf{Average} \\ \midrule
M-JAVE (2020)* & Extractive & 0.00 & 0.00 & 0.00 & 0.00 & 0.00 & 0.00 & 0.00 & 0.00 & 0.00 & 0.00 \\
CMA-CLIP (2022) & Discriminative & 5.92 & 14.52 & 29.08 & 11.60 & 22.02 & 45.68 & 13.31 & 27.54 & 49.56 & 24.36 \\
DEFLATE (2023) & Generative & 13.29 & 25.23 & 56.52 & 11.43 & 35.94 & 74.80 & 9.75 & 39.22 & 59.11 & 36.14 \\
EIVEN (Ours) & Generative & 34.92 & 61.21 & 74.61 & 38.80 & 74.44 & 84.20 & 32.27 & 64.98 & 76.31 & 60.19 \\ \midrule
Absolute Gains (\%p) & - & 21.63 & 35.98 & 18.09 & 27.37 & 38.50 & 9.40 & 22.52 & 25.76 & 17.20 & 24.05 \\ \bottomrule
\end{tabular}%
}
\caption{Performance (micro-F1) comparison with representative work across different approaches. Models are trained with 10, 100, all (up to 1000) labeled data per attribute value. EIVEN delivers best results on all datasets, surpassing the latest implicit attribute value extraction work DEFLATE \cite{zhang-etal-2023-pay} by 24.05\textit{\%p} on average. *Extractive approaches such as M-JAVE \cite{zhu-etal-2020-multimodal} fail to handle implicit attribute values that do not appear explicitly as a subsequence of product text.}
\label{tab:main_result_table}
\end{table*}

\begin{figure*}[!thb]
    \centering
    \includegraphics[width=1\textwidth]{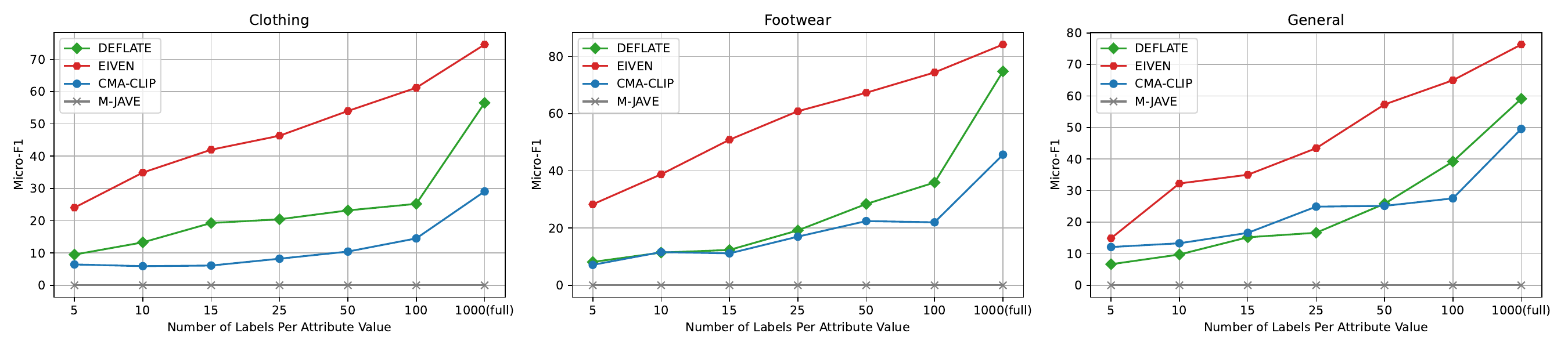} 
    \caption{Data efficiency demonstration with varying numbers of labeled data. EIVEN can achieve better performance than DEFLATE with less labeled data, highlighting its data efficiency.}
    \label{fig:main_result_figure}
\end{figure*}

\subsection{Learning-by-Comparison}

Many attributes have very similar attribute values, such as `Crew Neck', `Scoop Neck', and `Cowl Neck', which can confuse models. To help models better distinguish these similar attribute values, we propose a new technique called Learning-by-Comparison (LBC) to assist model training.

During training, in addition to the original product information $I_1, C_1$, and the query attribute $A$, we randomly sample another product with the same attribute $A$ and include its image $I_2$ and text context $C_2$ in the model input for comparison. We have designed three strategies: LBC\_Judge\_Last, LBC\_Judge\_First, and LBC\_Better\_Instance as illustrated in \textcolor{red}{Figure \ref{fig:LBC}}. We modify the attribute-related question and ground-truth answer accordingly. For example, in LBC\_Judge\_Last, we first ask the model to identify the value of the query attribute for both products, and then ask the model to compare and determine whether they have the same attribute value. The answer should be in the format of "First: \{attribute value of the first product\}; Second: \{attribute value of the second product\}; \{comparison result\}". Through this approach, the model is compelled to distinguish similar attribute values. Note that during the validation and testing phase, only the original product information and the attribute-related question are used.


\begin{table*}[tbh]
\centering
\resizebox{0.9\textwidth}{!}{%
\begin{tabular}{@{}lccccccccccc@{}}
\toprule
 &
  \multicolumn{1}{l}{} &
  \multicolumn{1}{l}{} &
  \multicolumn{1}{l}{} &
  \multicolumn{1}{l}{} &
  \multicolumn{2}{c}{\textbf{Clothing}} &
  \multicolumn{2}{c}{\textbf{Footwear}} &
  \multicolumn{2}{c}{\textbf{General}} &
  \multicolumn{1}{l}{} \\ \midrule
\textbf{Methods}        & \textbf{MGVF} & \textbf{LBC} & \textbf{Image} & \textbf{Text} & 50    & 100   & 50    & 100   & 50    & 100   & \textbf{Average} \\ \midrule
EIVEN          & \textcolor{teal}{\cmark}    & \textcolor{teal}{\cmark}   & \textcolor{teal}{\cmark}     & \textcolor{teal}{\cmark}            & 54.01 & 61.21 & 67.33 & 74.44 & 57.31 & 64.98 & 63.21   \\
- MGVF         & \textcolor{purple}{\xmark}    & \textcolor{teal}{\cmark}   & \textcolor{teal}{\cmark}     & \textcolor{teal}{\cmark}            & 49.92 & 57.75 & 65.04 & 72.73 & 53.5  & 62.27 & 60.20   \\
EIVEN-Base     & \textcolor{purple}{\xmark}    & \textcolor{purple}{\xmark}   & \textcolor{teal}{\cmark}     & \textcolor{teal}{\cmark}            & 49.76 & 55.50 & 64.14 & 73.46 & 47.85 & 59.30 & 58.34   \\
- Image        & \textcolor{purple}{\xmark}    & \textcolor{purple}{\xmark}   & \textcolor{purple}{\xmark}     & \textcolor{teal}{\cmark}            & 43.97 & 50.45 & 54.72 & 68.01 & 37.20 & 49.40 & 50.63   \\
- Text Context & \textcolor{purple}{\xmark}    & \textcolor{purple}{\xmark}   & \textcolor{teal}{\cmark}     & \textcolor{purple}{\xmark}          & 16.49 & 19.91 & 22.25 & 29.38 & 11.96 & 18.28 & 19.71   \\ \bottomrule
\end{tabular}%
}
\caption{Ablation study of key components and modality information. '50/100' represents the number of labels per attribute value, as is the case for the subsequent tables. "MGVF" denotes multi-granularity visual features.}
\label{tab:ablation_components}
\end{table*}


\section{Open-Source Multimodal Implicit AVE Dataset}  
\label{sec:dataset}

Multimodal implicit AVE is an emerging problem, and there is currently a lack of truly open-sourced datasets for multimodal implicit AVE. \footnote{The claimed released multimodal implicit AVE dataset from DEFLATE \cite{zhang-etal-2023-pay} is encrypted, and our multiple attempts to request decrypted data have failed.} Existing AVE datasets either do not contain product images or lack implicit attribute values. Thus, in this section, we introduce and make available several datasets to facilitate further research in this area.

Specifically, we present three multimodal implicit AVE datasets: Clothing, Footwear, and General. The statistics of these datasets are summarized in \textcolor{red}{Table \ref{tab:dataset_statistics}}. All of them are derived and sampled from two publicly available datasets, MAVE \cite{yang2022mave} and Amazon Reviews 2018 \cite{ni-etal-2019-justifying}. There are a total of 68,423 samples that cover 12 diverse product attributes and 87 common attribute values. Specifically, for each product attribute, we randomly collect product instances including the product texts (titles and product categories) and attribute values from the MAVE dataset. We collect popular attribute values with more than 100 instances for effective evaluation and randomly sample up to 1000 instances per attribute value to limit the dataset size. Since the MAVE dataset does not provide product images and is derived from the multimodal Amazon Reviews 2018 dataset, we collect the corresponding product images from the Amazon Reviews 2018 dataset using their shared product identification number. Furthermore, the MAVE dataset contains only explicit attribute values. To evaluate performance on implicit attribute value extraction, we manually removed all explicit attribute value mentions from the product text for each product and its corresponding attribute. Therefore, attribute values in these data can only be inferred from product images, text context, or prior knowledge, i.e., implicit attribute values. Lastly, we split the train, test, and validation sets in a ratio of 0.75:0.15:0.15. We open-sourced these datasets.

\section{Experiment}

\subsection{Experimental Setup}

\textbf{Baselines:} We compare EIVEN with representative baselines in multimodal AVE: the latest generative work DEFLATE \cite{zhang-etal-2023-pay}, the representative discriminative work CMA-CLIP \cite{9897323} and the extractive work M-JAVE \cite{zhu-etal-2020-multimodal}. Detailed descriptions of baselines are provided in Appendix \ref{sec:baseline_descrp}. \textbf{Metrics:} Following the latest work \cite{zhang-etal-2023-pay}, micro-F1 (\%) is used as our evaluation metric and we determine whether the extraction results are correct using the exact match criteria, in which the full sequence of words is required to be correct. 



\vspace{0.2cm}
\noindent \textbf{Implementation Details:} We select the ViT-B/16 \cite{dosovitskiy2021an} of the pre-trained CLIP \cite{radford2021learning} as our image encoder. The multi-granularity visual features contain 4 $[cls]$ tokens extracted from every 3 layer of ViT-B/16. We use LLaMA-7B \cite{Touvron2023LLaMAOA} as our LLM. The default dimension of the two-layer visual projection network is set to 128, and the dimension of the adapter in LLM is set to 8. LBC\_Judge\_Last is used as our default Learning-by-Comparison strategy. RepAdapter \cite{Luo2023TowardsEV, luo2023cheap} is adopted as our LLM adapter in default. We use AdamW \cite{loshchilov2018decoupled} as the optimizer and train the model for 15 epochs. During the generation stage, we use top-p sampling as our decoding strategy with the temperature of 0.1 and the top-p value of 0.75. We report the micro-F1 result from a single run.


\subsection{Performance Comparison with Baselines}
The micro-F1 results with varying numbers of labeled data on the three multimodal datasets are shown in \textcolor{red}{Table \ref{tab:main_result_table}} and \textcolor{red}{Figure \ref{fig:main_result_figure}}. As can be seen from these comparison results, EIVEN can deliver significantly better performance on average than the other baseline methods. For instance, EIVEN can surpass the recent generative approach DEFLATE by 18.09\textit{\%p} on the Clothing dataset and 17.20\textit{\%p} on the General dataset. Also, EIVEN is much more data-efficient compared to previous generative attribute value extraction approaches. Using only 100 labels per attribute value, EIVEN can outperform or perform on par with other baselines trained with all labels (i.e., 1000 labels per attribute value) on all three datasets. These results indicate the effectiveness of our efficient multimodal LLM framework with the Learning-by-Comparison technique. 



\begin{table}[!t]
\resizebox{\columnwidth}{!}{%
\begin{tabular}{@{}lccccccc@{}}
\toprule
 & \multicolumn{2}{c}{\textbf{Clothing}} & \multicolumn{2}{c}{\textbf{Footwear}} & \multicolumn{2}{c}{\textbf{General}} & \multicolumn{1}{l}{} \\ \midrule
 \textbf{Methods}                    & 50    & 100   & 50    & 100   & 50    & 100   & \textbf{Average} \\ \midrule
LBC\_Judge\_Last      & 54.01 & 61.21 & 67.33 & 74.44 & 57.31 & 64.98 & 63.21   \\
LBC\_Judge\_First     & 53.08 & 60.25 & 66.78 & 74.64 & 54.97 & 64.71 & 62.41   \\
LBC\_Better\_Instance & 52.34 & 60.22 & 68.26 & 73.51 & 53.02 & 63.51 & 61.81   \\
w/o LBC               & 49.76 &	55.50 & 64.14 &	73.46 &	47.85 &	59.30 & 58.34   \\ \bottomrule
\end{tabular}%
}
\caption{Ablation study on Learning-by-Comparison (LBC) strategies. All three strategies help improve performance, indicating their effectiveness in reducing model confusion. A visualization of the confusion matrix is provided in Appendix \ref{sec:conf_matrix}.}
\label{tab:ablation_lbc}
\end{table}


\begin{figure*}[!t]
    \centering
    \includegraphics[width=1\textwidth]{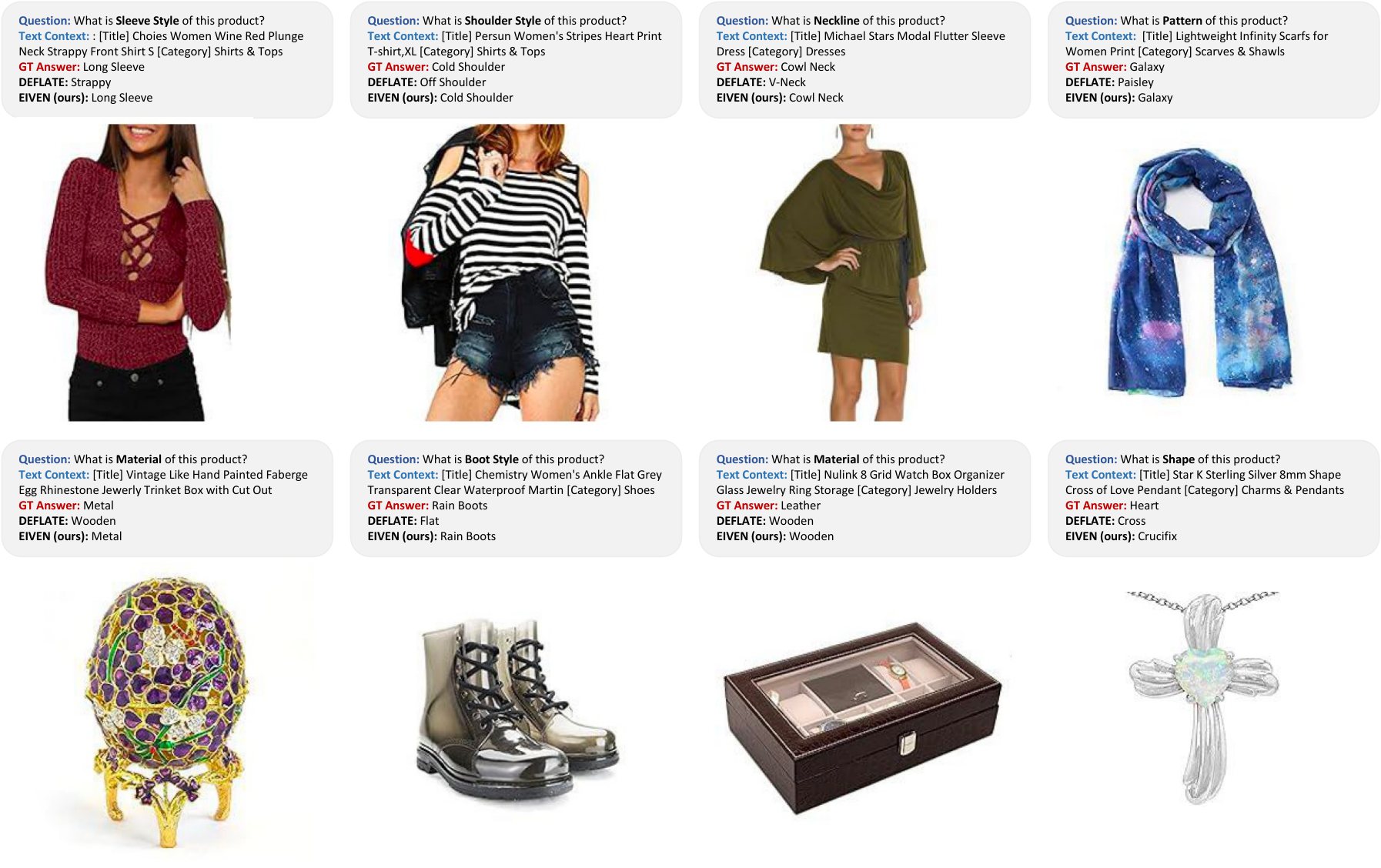} 
    \caption{Qualitative examples and comparisons between EIVEN and DEFLATE.}
    \label{fig:qualitative_example}
\end{figure*}

\section{Ablation Study and Analysis}

\subsection{Effectiveness of Each Component}
In order to quantify the impact of each component and modality in EIVEN, we measure and summarize the micro-F1 result of EIVEN after removing different components and modalities in \textcolor{red}{Table \ref{tab:ablation_components}}. First, we observe that the performance decreases after replacing multi-granularity visual features with the single-granularity feature 
or removing Learning-by-Comparison, suggesting that both of them contribute to the final performance of EIVEN. Notably, the performance of EIVEN-Base is still much better than DEFLATE, justifying the significant benefits of leveraging the LLM for implicit AVE. Besides, we can see that removing either the image or text context can significantly hurt model performance, which demonstrates the necessity of combining all these modalities in the implicit attribute value extraction task. Interestingly, the text modality plays the most important role, even when most of the ground truth attribute values cannot be explicitly identified from the product text. The possible reason is that implicit attribute values can still be inferred from the text context given the strong prior knowledge learned in LLM, as illustrated in the second product in \textcolor{red}{Figure \ref{fig:cover_example}}. On the other hand, extracting some product attribute values from images requires fine-grained visual understanding and thus is more challenging, especially when labels are limited.

\subsection{Learning-by-Comparison Strategies}

We explore different Learning-by-Comparison (LBC) strategies as illustrated in \textcolor{red}{Figure \ref{fig:LBC}}. The results of these strategies are presented in \textcolor{red}{Table \ref{tab:ablation_lbc}}. It is evident that all three strategies help improve the model's performance. This validates our motivation that including two instances into the model's input and asking the model to compare their attribute values can help alleviate model confusion among similar attribute values and improve overall performance. While there is no significant difference in performance among the three strategies, we believe that more effective LBC strategies can be devised to further enhance the model's performance, and we leave them for future exploration.

\subsection{Qualitative Examples}
\label{sec:qualitative_compare}

\textcolor{red}{Figure \ref{fig:qualitative_example}} demonstrates diverse qualitative examples and responses from the most recent generative work in implicit attribute value extraction DEFLATE and our method EIVEN. Compared to DEFLATE, EIVEN achieves overall better generation results across diverse product categories and attributes. In the first example, EIVEN extracts the correct attribute values for the product's sleeve style from the product image. In contrast, DEFLATE is confused by the strap in the neckline and generates incorrect answers. In the sixth example, EIVEN demonstrates its ability to infer the correct value "Rain Boots" for the attribute "Boot Style" from the text context "Transparent Clear Waterproof Martin", prior knowledge, and product image. We also visualize some failure cases in the last two examples. We observe that EIVEN can make mistakes when multiple reasonable attribute values exist.

\vspace{2pt}


\section{Conclusion}

In this paper, we propose EIVEN, an efficient generative framework using multimodal LLM for implicit attribute value extraction. EIVEN leverages the rich internal knowledge of pre-trained LLM to reduce reliance on attribute-specific labeled data and adopts lightweight adapters for parameter-efficient fine-tuning of LLM. Besides, to enhance the visual understanding ability of our model, we feed multi-granularity visual features into LLM and propose Learning-by-Comparison strategies to alleviate model confusion among attribute values. We also release the first open-source dataset. Through extensive experiments on three multimodal implicit attribute value extraction datasets, we found that EIVEN can significantly outperform previous works using fewer labels, making it an efficient solution for implicit attribute value extraction.

\section*{Limitations}

There are several limitations to our work. First, we only compared our approach with a limited number of baselines. This is because implicit multimodal attribute value extraction is a relatively new task, and also most of other multimodal attribute value extraction works are not open-sourced and very difficult to reproduce. We are planning to establish the first open-source benchmark for multimodal implicit AVE, which will also include comparisons among pre-trained general-purpose multimodal LLMs such as InstructBLIP \cite{dai2023instructblip}, LLaVA \cite{liu2023visual} and GPT-4V. Second, we observed that some annotations from MAVE \cite{yang2022mave} are not accurate for implicit attribute value extraction, and there are some semantically overlapping attribute values. Automatic correction methods and human inspections are needed to construct more suitable benchmark datasets for implicit attribute value extraction. We plan to conduct such exploration in the future. In addition, more effective LBC strategies can be devised to further improve model performance.

\bibliography{anthology,custom}

\begin{thebibliography}{39}
\expandafter\ifx\csname natexlab\endcsname\relax\def\natexlab#1{#1}\fi

\bibitem[{Blume et~al.(2023)Blume, Zalmout, Ji, and Li}]{blume-etal-2023-generative}
Ansel Blume, Nasser Zalmout, Heng Ji, and Xian Li. 2023.
\newblock \href {https://aclanthology.org/2023.emnlp-industry.55} {Generative models for product attribute extraction}.
\newblock In \emph{Proceedings of the 2023 Conference on Empirical Methods in Natural Language Processing: Industry Track}, pages 575--585, Singapore. Association for Computational Linguistics.

\bibitem[{Chen et~al.(2022)Chen, Xia, and Shinzato}]{chen-etal-2022-extreme}
Wei-Te Chen, Yandi Xia, and Keiji Shinzato. 2022.
\newblock \href {https://doi.org/10.18653/v1/2022.ecnlp-1.16} {Extreme multi-label classification with label masking for product attribute value extraction}.
\newblock In \emph{Proceedings of the Fifth Workshop on e-Commerce and NLP (ECNLP 5)}, pages 134--140, Dublin, Ireland. Association for Computational Linguistics.

\bibitem[{Dai et~al.(2023)Dai, Li, Li, Tiong, Zhao, Wang, Li, Fung, and Hoi}]{dai2023instructblip}
Wenliang Dai, Junnan Li, Dongxu Li, Anthony Tiong, Junqi Zhao, Weisheng Wang, Boyang Li, Pascale Fung, and Steven Hoi. 2023.
\newblock \href {https://openreview.net/forum?id=vvoWPYqZJA} {Instruct{BLIP}: Towards general-purpose vision-language models with instruction tuning}.
\newblock In \emph{Thirty-seventh Conference on Neural Information Processing Systems}.

\bibitem[{Dong et~al.(2023)Dong, Chen, Liu, Polak, and Zhang}]{Dong2023MuseChatAC}
Zhikang Dong, Bin Chen, Xiulong Liu, Pawel Polak, and Peng Zhang. 2023.
\newblock \href {https://api.semanticscholar.org/CorpusID:263830510} {Musechat: A conversational music recommendation system for videos}.
\newblock \emph{ArXiv}, abs/2310.06282.

\bibitem[{Dosovitskiy et~al.(2021)Dosovitskiy, Beyer, Kolesnikov, Weissenborn, Zhai, Unterthiner, Dehghani, Minderer, Heigold, Gelly, Uszkoreit, and Houlsby}]{dosovitskiy2021an}
Alexey Dosovitskiy, Lucas Beyer, Alexander Kolesnikov, Dirk Weissenborn, Xiaohua Zhai, Thomas Unterthiner, Mostafa Dehghani, Matthias Minderer, Georg Heigold, Sylvain Gelly, Jakob Uszkoreit, and Neil Houlsby. 2021.
\newblock \href {https://openreview.net/forum?id=YicbFdNTTy} {An image is worth 16x16 words: Transformers for image recognition at scale}.
\newblock In \emph{International Conference on Learning Representations}.

\bibitem[{Fu et~al.(2022)Fu, Xu, Liu, Liu, Xie, Wang, Liu, Sun, and Wang}]{9897323}
Jinmiao Fu, Shaoyuan Xu, Huidong Liu, Yang Liu, Ning Xie, Chien-Chih Wang, Jia Liu, Yi~Sun, and Bryan Wang. 2022.
\newblock \href {https://doi.org/10.1109/ICIP46576.2022.9897323} {Cma-clip: Cross-modality attention clip for text-image classification}.
\newblock In \emph{2022 IEEE International Conference on Image Processing (ICIP)}, pages 2846--2850.

\bibitem[{Ghiasi et~al.(2022)Ghiasi, Kazemi, Borgnia, Reich, Shu, Goldblum, Wilson, and Goldstein}]{Ghiasi2022WhatDV}
Amin Ghiasi, Hamid Kazemi, Eitan Borgnia, Steven Reich, Manli Shu, Micah Goldblum, Andrew~Gordon Wilson, and Tom Goldstein. 2022.
\newblock \href {https://api.semanticscholar.org/CorpusID:254591270} {What do vision transformers learn? a visual exploration}.
\newblock \emph{ArXiv}, abs/2212.06727.

\bibitem[{Houlsby et~al.(2019)Houlsby, Giurgiu, Jastrzebski, Morrone, De~Laroussilhe, Gesmundo, Attariyan, and Gelly}]{houlsby2019parameter}
Neil Houlsby, Andrei Giurgiu, Stanislaw Jastrzebski, Bruna Morrone, Quentin De~Laroussilhe, Andrea Gesmundo, Mona Attariyan, and Sylvain Gelly. 2019.
\newblock \href {https://api.semanticscholar.org/CorpusID:59599816} {Parameter-efficient transfer learning for nlp}.
\newblock In \emph{International Conference on Machine Learning}, pages 2790--2799. PMLR.

\bibitem[{Hu et~al.(2023)Hu, Wang, Lan, Xu, Lim, Bing, Xu, Poria, and Lee}]{hu-etal-2023-llm}
Zhiqiang Hu, Lei Wang, Yihuai Lan, Wanyu Xu, Ee-Peng Lim, Lidong Bing, Xing Xu, Soujanya Poria, and Roy Lee. 2023.
\newblock \href {https://aclanthology.org/2023.emnlp-main.319} {{LLM}-adapters: An adapter family for parameter-efficient fine-tuning of large language models}.
\newblock In \emph{Proceedings of the 2023 Conference on Empirical Methods in Natural Language Processing}, pages 5254--5276, Singapore. Association for Computational Linguistics.

\bibitem[{Khandelwal et~al.(2023)Khandelwal, Mittal, Kulkarni, and Gupta}]{khandelwal-etal-2023-large}
Anant Khandelwal, Happy Mittal, Shreyas Kulkarni, and Deepak Gupta. 2023.
\newblock \href {https://doi.org/10.18653/v1/2023.acl-industry.29} {Large scale generative multimodal attribute extraction for {E}-commerce attributes}.
\newblock In \emph{Proceedings of the 61st Annual Meeting of the Association for Computational Linguistics (Volume 5: Industry Track)}, pages 305--312, Toronto, Canada. Association for Computational Linguistics.

\bibitem[{Lai et~al.(2024)Lai, Bai, Zhang, Du, Shan, Yang, Chuah, and Cao}]{lai2024empowering}
Zhengfeng Lai, Haoping Bai, Haotian Zhang, Xianzhi Du, Jiulong Shan, Yinfei Yang, Chen-Nee Chuah, and Meng Cao. 2024.
\newblock \href {https://api.semanticscholar.org/CorpusID:268753696} {Empowering unsupervised domain adaptation with large-scale pre-trained vision-language models}.
\newblock In \emph{Proceedings of the IEEE/CVF Winter Conference on Applications of Computer Vision}, pages 2691--2701.

\bibitem[{Lai et~al.(2021)Lai, Wang, Hu, Dugger, Cheung, and Chuah}]{Lai2021ASL}
Zhengfeng Lai, Chao Wang, Zin Hu, Brittany~N. Dugger, Sen-Ching~Samson Cheung, and Chen-Nee Chuah. 2021.
\newblock \href {https://api.semanticscholar.org/CorpusID:243867193} {A semi-supervised learning for segmentation of gigapixel histopathology images from brain tissues}.
\newblock \emph{2021 43rd Annual International Conference of the IEEE Engineering in Medicine \& Biology Society (EMBC)}, pages 1920--1923.

\bibitem[{Li et~al.(2023)Li, Xue, Zhang, and Zou}]{li-etal-2023-attgen}
Yanzeng Li, Bingcong Xue, Ruoyu Zhang, and Lei Zou. 2023.
\newblock \href {https://doi.org/10.18653/v1/2023.acl-long.119} {{A}t{TG}en: Attribute tree generation for real-world attribute joint extraction}.
\newblock In \emph{Proceedings of the 61st Annual Meeting of the Association for Computational Linguistics (Volume 1: Long Papers)}, pages 2139--2152, Toronto, Canada. Association for Computational Linguistics.

\bibitem[{Lin et~al.(2021)Lin, He, Feng, Zalmout, Liang, Xiong, and Dong}]{lin2021pam}
Rongmei Lin, Xiang He, Jie Feng, Nasser Zalmout, Yan Liang, Li~Xiong, and Xin~Luna Dong. 2021.
\newblock \href {https://dl.acm.org/doi/10.1145/3447548.3467164} {Pam: Understanding product images in cross product category attribute extraction}.
\newblock In \emph{Proceedings of the 27th ACM SIGKDD Conference on Knowledge Discovery \& Data Mining}, pages 3262--3270.

\bibitem[{Liu et~al.(2023)Liu, Li, Wu, and Lee}]{liu2023visual}
Haotian Liu, Chunyuan Li, Qingyang Wu, and Yong~Jae Lee. 2023.
\newblock \href {https://openreview.net/forum?id=w0H2xGHlkw} {Visual instruction tuning}.
\newblock In \emph{Thirty-seventh Conference on Neural Information Processing Systems}.

\bibitem[{Loshchilov and Hutter(2019)}]{loshchilov2018decoupled}
Ilya Loshchilov and Frank Hutter. 2019.
\newblock \href {https://openreview.net/forum?id=Bkg6RiCqY7} {Decoupled weight decay regularization}.
\newblock In \emph{International Conference on Learning Representations}.

\bibitem[{Luo et~al.(2023{\natexlab{a}})Luo, Huang, Zhou, Sun, Jiang, Wang, and Ji}]{Luo2023TowardsEV}
Gen Luo, Minglang Huang, Yiyi Zhou, Xiaoshuai Sun, Guannan Jiang, Zhiyu Wang, and Rongrong Ji. 2023{\natexlab{a}}.
\newblock \href {https://api.semanticscholar.org/CorpusID:256900990} {Towards efficient visual adaption via structural re-parameterization}.
\newblock \emph{ArXiv}, abs/2302.08106.

\bibitem[{Luo et~al.(2023{\natexlab{b}})Luo, Zhou, Ren, Chen, Sun, and Ji}]{luo2023cheap}
Gen Luo, Yiyi Zhou, Tianhe Ren, Shengxin Chen, Xiaoshuai Sun, and Rongrong Ji. 2023{\natexlab{b}}.
\newblock \href {https://openreview.net/forum?id=t877958UGZ} {Cheap and quick: Efficient vision-language instruction tuning for large language models}.
\newblock In \emph{Thirty-seventh Conference on Neural Information Processing Systems}.

\bibitem[{Nguyen et~al.(2019)Nguyen, Yosinski, and Clune}]{nguyen2019understanding}
Anh Nguyen, Jason Yosinski, and Jeff Clune. 2019.
\newblock \href {https://api.semanticscholar.org/CorpusID:125954000} {Understanding neural networks via feature visualization: A survey}.
\newblock \emph{Explainable AI: interpreting, explaining and visualizing deep learning}, pages 55--76.

\bibitem[{Ni et~al.(2019)Ni, Li, and McAuley}]{ni-etal-2019-justifying}
Jianmo Ni, Jiacheng Li, and Julian McAuley. 2019.
\newblock \href {https://doi.org/10.18653/v1/D19-1018} {Justifying recommendations using distantly-labeled reviews and fine-grained aspects}.
\newblock In \emph{Proceedings of the 2019 Conference on Empirical Methods in Natural Language Processing and the 9th International Joint Conference on Natural Language Processing (EMNLP-IJCNLP)}, pages 188--197, Hong Kong, China. Association for Computational Linguistics.

\bibitem[{Radford et~al.(2021)Radford, Kim, Hallacy, Ramesh, Goh, Agarwal, Sastry, Askell, Mishkin, Clark et~al.}]{radford2021learning}
Alec Radford, Jong~Wook Kim, Chris Hallacy, Aditya Ramesh, Gabriel Goh, Sandhini Agarwal, Girish Sastry, Amanda Askell, Pamela Mishkin, Jack Clark, et~al. 2021.
\newblock \href {https://proceedings.mlr.press/v139/radford21a/radford21a.pdf} {Learning transferable visual models from natural language supervision}.
\newblock In \emph{International conference on machine learning}, pages 8748--8763. PMLR.

\bibitem[{Shazeer(2020)}]{shazeer2020glu}
Noam~M. Shazeer. 2020.
\newblock \href {https://api.semanticscholar.org/CorpusID:211096588} {Glu variants improve transformer}.
\newblock \emph{ArXiv}, abs/2002.05202.

\bibitem[{Shinzato et~al.(2023)Shinzato, Yoshinaga, Xia, and Chen}]{shinzato-etal-2023-unified}
Keiji Shinzato, Naoki Yoshinaga, Yandi Xia, and Wei-Te Chen. 2023.
\newblock \href {https://doi.org/10.18653/v1/2023.findings-acl.413} {A unified generative approach to product attribute-value identification}.
\newblock In \emph{Findings of the Association for Computational Linguistics: ACL 2023}, pages 6599--6612, Toronto, Canada. Association for Computational Linguistics.

\bibitem[{Tian et~al.(2024)Tian, Han, Chen, Wang, and Chawla}]{Tian2024TinyLLMLA}
Yijun Tian, Yikun Han, Xiusi Chen, Wei Wang, and N.~Chawla. 2024.
\newblock \href {https://api.semanticscholar.org/CorpusID:267523447} {Tinyllm: Learning a small student from multiple large language models}.
\newblock \emph{ArXiv}, abs/2402.04616.

\bibitem[{Tian et~al.(2023)Tian, Song, Wang, Wang, Hu, Wang, Chawla, and Xu}]{Tian2023GraphNP}
Yijun Tian, Huan Song, Zichen Wang, Haozhu Wang, Ziqing Hu, Fang Wang, N.~Chawla, and Panpan Xu. 2023.
\newblock \href {https://api.semanticscholar.org/CorpusID:263152125} {Graph neural prompting with large language models}.
\newblock In \emph{AAAI Conference on Artificial Intelligence}.

\bibitem[{Touvron et~al.(2023)Touvron, Lavril, Izacard, Martinet, Lachaux, Lacroix, Rozi{\`e}re, Goyal, Hambro, Azhar, Rodriguez, Joulin, Grave, and Lample}]{Touvron2023LLaMAOA}
Hugo Touvron, Thibaut Lavril, Gautier Izacard, Xavier Martinet, Marie-Anne Lachaux, Timoth{\'e}e Lacroix, Baptiste Rozi{\`e}re, Naman Goyal, Eric Hambro, Faisal Azhar, Aurelien Rodriguez, Armand Joulin, Edouard Grave, and Guillaume Lample. 2023.
\newblock \href {https://api.semanticscholar.org/CorpusID:257219404} {Llama: Open and efficient foundation language models}.
\newblock \emph{ArXiv}, abs/2302.13971.

\bibitem[{Wang et~al.(2020)Wang, Yang, Kanagal, Sanghai, Sivakumar, Shu, Yu, and Elsas}]{wang2020learning}
Qifan Wang, Li~Yang, Bhargav Kanagal, Sumit Sanghai, D~Sivakumar, Bin Shu, Zac Yu, and Jon Elsas. 2020.
\newblock \href {https://dl.acm.org/doi/pdf/10.1145/3394486.3403047} {Learning to extract attribute value from product via question answering: A multi-task approach}.
\newblock In \emph{Proceedings of the 26th ACM SIGKDD international conference on knowledge discovery \& data mining}, pages 47--55.

\bibitem[{Wang et~al.(2022)Wang, Yang, Wang, Krishnan, Dai, Wang, Xu, Khabsa, and Ma}]{wang-etal-2022-smartave}
Qifan Wang, Li~Yang, Jingang Wang, Jitin Krishnan, Bo~Dai, Sinong Wang, Zenglin Xu, Madian Khabsa, and Hao Ma. 2022.
\newblock \href {https://doi.org/10.18653/v1/2022.findings-emnlp.20} {{SMARTAVE}: Structured multimodal transformer for product attribute value extraction}.
\newblock In \emph{Findings of the Association for Computational Linguistics: EMNLP 2022}, pages 263--276, Abu Dhabi, United Arab Emirates. Association for Computational Linguistics.

\bibitem[{Wang et~al.(2023)Wang, Lipka, Rossi, Siu, Zhang, and Derr}]{Wang2023KnowledgeGP}
Yu~Wang, Nedim Lipka, Ryan~A. Rossi, Alexa~F. Siu, Ruiyi Zhang, and Tyler Derr. 2023.
\newblock \href {https://api.semanticscholar.org/CorpusID:261076072} {Knowledge graph prompting for multi-document question answering}.
\newblock In \emph{AAAI Conference on Artificial Intelligence}.

\bibitem[{Xu et~al.(2019)Xu, Wang, Mao, Jiang, and Lan}]{xu-etal-2019-scaling}
Huimin Xu, Wenting Wang, Xin Mao, Xinyu Jiang, and Man Lan. 2019.
\newblock \href {https://doi.org/10.18653/v1/P19-1514} {Scaling up open tagging from tens to thousands: Comprehension empowered attribute value extraction from product title}.
\newblock In \emph{Proceedings of the 57th Annual Meeting of the Association for Computational Linguistics}, pages 5214--5223, Florence, Italy. Association for Computational Linguistics.

\bibitem[{Xu et~al.(2023)Xu, Zhang, Li, Shang, and Choi}]{xu-etal-2023-towards}
Liyan Xu, Chenwei Zhang, Xian Li, Jingbo Shang, and Jinho~D. Choi. 2023.
\newblock \href {https://doi.org/10.18653/v1/2023.acl-long.683} {Towards open-world product attribute mining: A lightly-supervised approach}.
\newblock In \emph{Proceedings of the 61st Annual Meeting of the Association for Computational Linguistics (Volume 1: Long Papers)}, pages 12223--12239, Toronto, Canada. Association for Computational Linguistics.

\bibitem[{Yan et~al.(2021)Yan, Zalmout, Liang, Grant, Ren, and Dong}]{yan-etal-2021-adatag}
Jun Yan, Nasser Zalmout, Yan Liang, Christan Grant, Xiang Ren, and Xin~Luna Dong. 2021.
\newblock \href {https://doi.org/10.18653/v1/2021.acl-long.362} {{A}da{T}ag: Multi-attribute value extraction from product profiles with adaptive decoding}.
\newblock In \emph{Proceedings of the 59th Annual Meeting of the Association for Computational Linguistics and the 11th International Joint Conference on Natural Language Processing (Volume 1: Long Papers)}, pages 4694--4705, Online. Association for Computational Linguistics.

\bibitem[{Yang et~al.(2023)Yang, Wang, Wang, Quan, Feng, Chen, Khabsa, Wang, Xu, and Liu}]{yang-etal-2023-mixpave}
Li~Yang, Qifan Wang, Jingang Wang, Xiaojun Quan, Fuli Feng, Yu~Chen, Madian Khabsa, Sinong Wang, Zenglin Xu, and Dongfang Liu. 2023.
\newblock \href {https://doi.org/10.18653/v1/2023.findings-acl.633} {{M}ix{PAVE}: Mix-prompt tuning for few-shot product attribute value extraction}.
\newblock In \emph{Findings of the Association for Computational Linguistics: ACL 2023}, pages 9978--9991, Toronto, Canada. Association for Computational Linguistics.

\bibitem[{Yang et~al.(2022)Yang, Wang, Yu, Kulkarni, Sanghai, Shu, Elsas, and Kanagal}]{yang2022mave}
Li~Yang, Qifan Wang, Zac Yu, Anand Kulkarni, Sumit Sanghai, Bin Shu, Jon Elsas, and Bhargav Kanagal. 2022.
\newblock \href {https://dl.acm.org/doi/10.1145/3488560.3498377} {Mave: A product dataset for multi-source attribute value extraction}.
\newblock In \emph{Proceedings of the fifteenth ACM international conference on web search and data mining}, pages 1256--1265.

\bibitem[{Zhang et~al.(2023)Zhang, Wang, Li, Dong, Wang, Xian, Li, and Zhang}]{zhang-etal-2023-pay}
Yupeng Zhang, Shensi Wang, Peiguang Li, Guanting Dong, Sirui Wang, Yunsen Xian, Zhoujun Li, and Hongzhi Zhang. 2023.
\newblock \href {https://doi.org/10.18653/v1/2023.findings-acl.831} {Pay attention to implicit attribute values: A multi-modal generative framework for {AVE} task}.
\newblock In \emph{Findings of the Association for Computational Linguistics: ACL 2023}, pages 13139--13151, Toronto, Canada. Association for Computational Linguistics.

\bibitem[{Zheng et~al.(2018)Zheng, Mukherjee, Dong, and Li}]{zheng2018opentag}
Guineng Zheng, Subhabrata Mukherjee, Xin~Luna Dong, and Feifei Li. 2018.
\newblock \href {https://dl.acm.org/doi/abs/10.1145/3219819.3219839} {Opentag: Open attribute value extraction from product profiles}.
\newblock In \emph{Proceedings of the 24th ACM SIGKDD international conference on knowledge discovery \& data mining}, pages 1049--1058.

\bibitem[{Zhu et~al.(2020)Zhu, Wang, Li, Wu, He, and Zhou}]{zhu-etal-2020-multimodal}
Tiangang Zhu, Yue Wang, Haoran Li, Youzheng Wu, Xiaodong He, and Bowen Zhou. 2020.
\newblock \href {https://doi.org/10.18653/v1/2020.emnlp-main.166} {Multimodal joint attribute prediction and value extraction for {E}-commerce product}.
\newblock In \emph{Proceedings of the 2020 Conference on Empirical Methods in Natural Language Processing (EMNLP)}, pages 2129--2139, Online. Association for Computational Linguistics.

\bibitem[{Zou and Caragea(2023)}]{zou-caragea-2023-jointmatch}
Henry Zou and Cornelia Caragea. 2023.
\newblock \href {https://doi.org/10.18653/v1/2023.emnlp-main.451} {{J}oint{M}atch: A unified approach for diverse and collaborative pseudo-labeling to semi-supervised text classification}.
\newblock In \emph{Proceedings of the 2023 Conference on Empirical Methods in Natural Language Processing}, pages 7290--7301, Singapore. Association for Computational Linguistics.

\bibitem[{Zou et~al.(2023)Zou, Zhou, Zhang, and Caragea}]{zou-etal-2023-decrisismb}
Henry Zou, Yue Zhou, Weizhi Zhang, and Cornelia Caragea. 2023.
\newblock \href {https://doi.org/10.18653/v1/2023.findings-emnlp.406} {{D}e{C}risis{MB}: Debiased semi-supervised learning for crisis tweet classification via memory bank}.
\newblock In \emph{Findings of the Association for Computational Linguistics: EMNLP 2023}, pages 6104--6115, Singapore. Association for Computational Linguistics.

\end{thebibliography}

\newpage

\appendix


\section{Detailed Discussion of Previous Works in Multimodal AVE}
\label{sec:discussion_multimodalAVE}

Existing approaches for multimodal attribute value extraction can be broadly categorized into three categories: extractive, discriminative, and generative (Table \ref{tab:compare_ave_approaches}). \textbf{Extractive} approaches pose this task as a named entity recognition or sequence tagging problem, where the model outputs the start and end positions of the attribute value in the input text \cite{zhu-etal-2020-multimodal, xu-etal-2019-scaling}. However, they are incapable of extracting implicit attribute values hidden in textual contexts or images. Additionally, they can only obtain raw value strings from product text, instead of the canonicalized values required for services such as faceted product search (e.g., 'Short Sleeve' instead of 'Short Sleeves' or 'Short Sleeved Shirt'). A further step is required for extractive approaches to canonicalize extracted raw value strings. \textbf{Discriminative} approaches classify each instance into a pre-defined set of attribute values \cite{9897323, chen-etal-2022-extreme}. Yet, they cannot identify attribute values not in the pre-defined set and are hard to scale to large amounts of attributes. Ideally, we would like to eliminate the need to re-train a separate model for every new attribute or attribute value. \textbf{Generative} approaches frame the task as generating answers to attribute-related queries, using product information as a reference \cite{lin2021pam, wang-etal-2022-smartave, khandelwal-etal-2023-large, zhang-etal-2023-pay}. Given their nature of free-form text output, they are able to address implicit attribute values, unseen values, and can learn to directly obtain canonicalized values and answer values for multiple attributes. Nonetheless, previous generative methods in multimodal attribute value extraction require large amounts of labeled data for training and still perform very poorly on datasets with implicit attribute values.

\begin{table}[!th]
\centering
\resizebox{\columnwidth}{!}{%
\begin{tabular}{@{}lcccc@{}}
\toprule
 \begin{tabular}[c]{@{}c@{}}\textbf{Approach}\end{tabular} & \begin{tabular}[c]{@{}c@{}}\textbf{Implicit} \\ \textbf{Values}\end{tabular} & \begin{tabular}[c]{@{}c@{}}\textbf{Unseen} \\ \textbf{Values}\end{tabular} & \begin{tabular}[c]{@{}c@{}}\textbf{Canonical} \\ \textbf{Values}\end{tabular} & \begin{tabular}[c]{@{}c@{}}\textbf{Scalable} \\ \textbf{Attributes}\end{tabular} \\ \midrule
Extractive & \textcolor{purple}{\xmark}  & \textcolor{teal}{\cmark} & \textcolor{purple}{\xmark} & \textcolor{teal}{\cmark} \\
Discriminative & \textcolor{teal}{\cmark} & \textcolor{purple}{\xmark} & \textcolor{teal}{\cmark} & \textcolor{purple}{\xmark} \\
Generative & \textcolor{teal}{\cmark} & \textcolor{teal}{\cmark} & \textcolor{teal}{\cmark} & \textcolor{teal}{\cmark} \\ \bottomrule
\end{tabular}%
}
\caption{Different AVE approaches and challenges.}
\label{tab:compare_ave_approaches}
\end{table}

\begin{table}[!ht]
\resizebox{\columnwidth}{!}{%
\begin{tabular}{@{}lcccccccc@{}}
\toprule
 & \multicolumn{1}{l}{} & \multicolumn{1}{l}{} & \multicolumn{1}{l}{} & \multicolumn{2}{c}{\textbf{Clothing}} & \multicolumn{2}{c}{\textbf{Footwear}} & \multicolumn{1}{l}{} \\ \midrule
 \textbf{Methods}  & \textbf{Linear} & \textbf{Sparse} & \textbf{\# Param} & 50 & 100 & 50 & 100 & \textbf{Average} \\ \midrule
RepAdapter & \textcolor{teal}{\cmark} & \textcolor{teal}{\cmark} & 1.70M & 49.76 & 55.50 & 64.14 & 73.46 & 60.72 \\
MLP-Adapter & \textcolor{purple}{\xmark} & \textcolor{purple}{\xmark} & 2.23M & 53.43 & 59.61 & 67.60 & 73.38 & 63.51 \\
MLP-Adapter-L & \textcolor{teal}{\cmark} & \textcolor{purple}{\xmark} & 2.23M & 45.86 & 54.89 & 64.92 & 69.81 & 58.87 \\ \bottomrule
\end{tabular}%
}
\caption{Ablation study on the adapter in EIVEN-Base.}
\label{tab:ablation_adapter}
\end{table}


\section{Dataset Statistics}

The statistics of the introduced multimodal implicit AVE datasets (Footwear, Clothing, General) are provided in Table \ref{tab:dataset_statistics}.

\begin{table*}[!tbh]
\centering
\resizebox{0.95\textwidth}{!}{%
\begin{tabular}{@{}cccccl@{}}
\toprule
\textbf{Dataset} & \textbf{\# Samples} & \textbf{\# Values} & \textbf{\# Head} & \textbf{\# Tail} & \multicolumn{1}{c}{\textbf{Attributes}} \\ \midrule
Footwear & 26868 & 32 & 1000 & 229 & Athletic Shoe Style, Boot Style, Shaft Height, Heel, Toe Style \\
Clothing & 24664 & 30 & 1000 & 211 & Neckline, Dress Length, Sleeve Style, Shoulder Style \\
General & 16891 & 25 & 1000 & 117 & Pattern, Material, Shape \\
Total & 68423 & 87 & 1000 & 117 & \multicolumn{1}{l}{-} \\ \bottomrule
\end{tabular}%
}
\caption{Dataset statistics. `'\# Head' and `'\# Tail' denote the maximum and minimum amounts of attribute value instances among all attributes in the dataset. More details about these datasets can be found in Section \ref{sec:dataset}.}
\label{tab:dataset_statistics}
\end{table*}

\section{Detailed Descriptions of Baselines}
\label{sec:baseline_descrp}

We describe in detail our baselines here: (1) \textbf{M-JAVE} \cite{zhu-etal-2020-multimodal}: A representative extractive approach that labels the input textual product description as "BIO" sequences related to attributes. It utilizes the fused multimodal features from the global and regional-gated cross-modality attention layer to make attribute predictions jointly. (2) \textbf{CMA-CLIP} \cite{9897323}: A recent discriminative approach that uses CLIP and sequence-wise attention to learn fine-grained multimodal product features. A modality-wise attention is then proposed to adaptively weigh the importance of visual and textual modalities to discriminate values for different product attributes. (3) \textbf{DEFLATE} \cite{zhang-etal-2023-pay}: A T5-based generative approach that consists of a generator to produce candidate attribute values from product information from different modalities and a discriminator to ensure the credibility of the generated answers.




\section{Ablation Study on Adapters}

\label{sec:ablation_adapters}

In this section, we study the performance of different types of adapters from the perspective of linearity and sparsity. RepAdapter \cite{Luo2023TowardsEV} is a recently proposed linear adapter without an activation function and has a sparse structure via group-wise transformation. The linear structure allows parameters in the adapter to be re-parameterized into LLM and thus introduces no inference latency. The sparse structure helps reduce the number of parameters and save memory consumption. Table \ref{tab:ablation_adapter} shows the comparison result with the representative MLP-adapter \cite{houlsby2019parameter} in LLM. MLP-Adapter performs the best in micro-F1, while RepAdapter has the fewest parameters. We also observe that the linear structure generally sacrifices model micro-F1 performance in our task, and sparse transformation can boost model performance as well as reduce the number of parameters.

\newpage
\section{Confusion Matrix}
\label{sec:conf_matrix}

\begin{figure}[!t]
     \centering
     \begin{subfigure}[b]{0.5\textwidth}
         \centering
         \includegraphics[width=\textwidth]{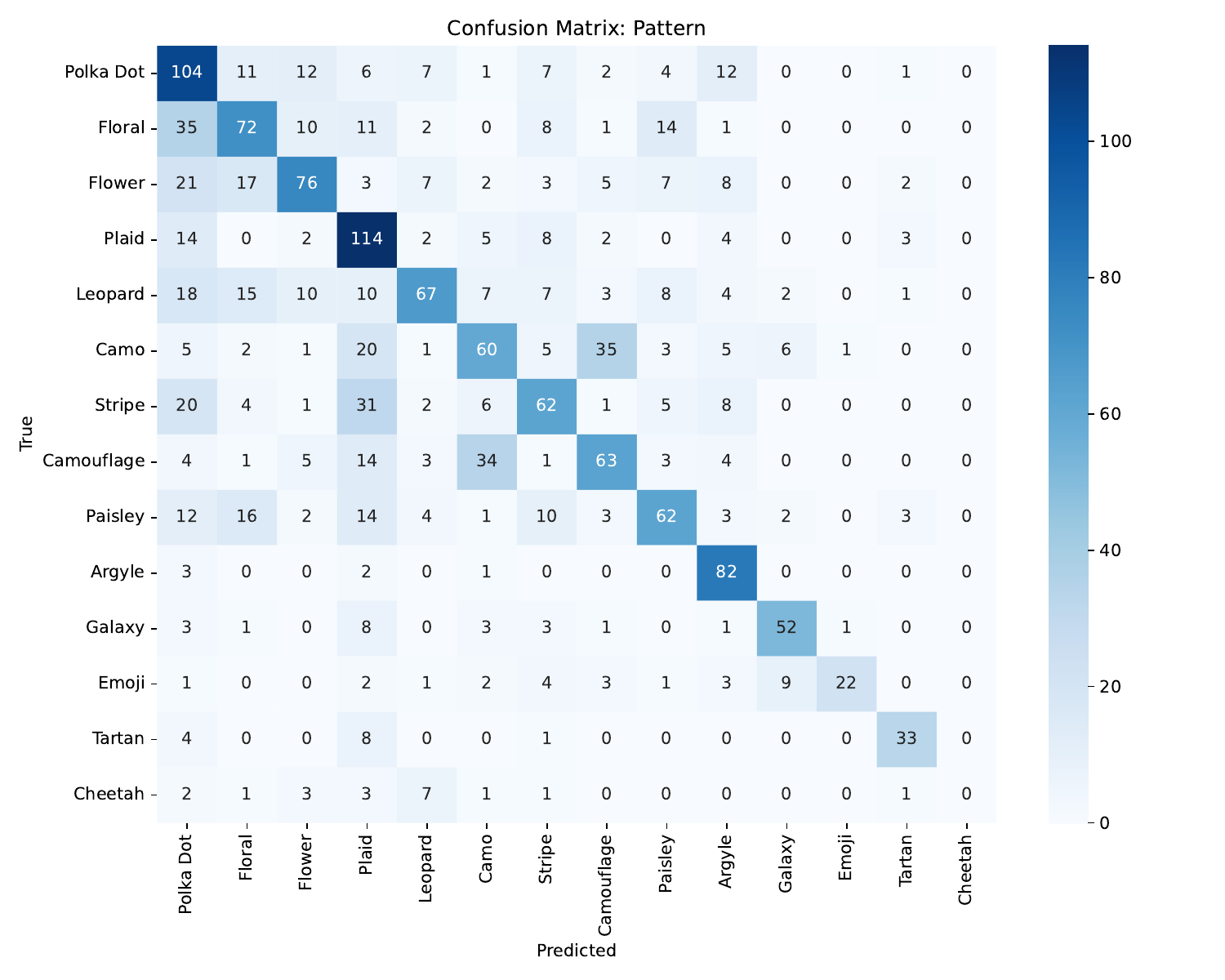}
         \caption{DEFLATE}
         \label{Figure 5(a)}
     \end{subfigure}
     \begin{subfigure}[b]{0.5\textwidth}
         \centering
         \includegraphics[width=\textwidth]{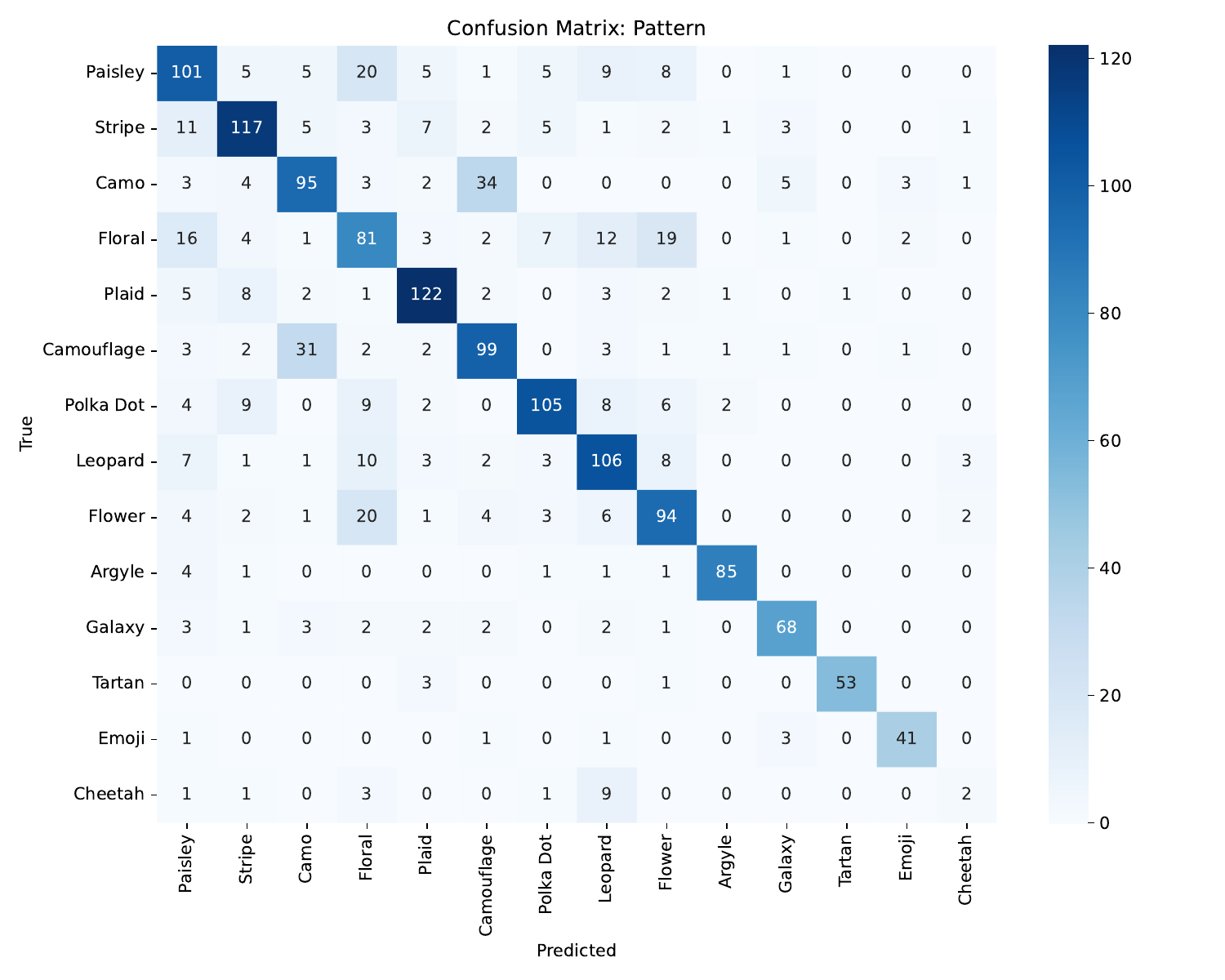}
         \caption{EIVEN}
         \label{Figure 5(b)}
     \end{subfigure}
        \caption{Confusion matrix for the Pattern attribute. LBC\_Judge\_Last is used in this example as the Learning-by-Comparison strategy. It can be observed that the confusion among attribute values is significantly reduced, demonstrating the effectiveness of our Learning-by-Comparison technique.}
        \label{confusion_matrix}
        
\end{figure}


Figure \ref{confusion_matrix} visualizes the confusion matrix of EIVEN and DEFLATE for the Pattern attribute on the General dataset using all labeled data. It can be observed that EIVEN has much less confusion compared to DEFLATE, which validates our utilization of LLM and our Learning-by-Comparison strategy.




\end{document}